\documentclass[journal]{IEEEtran}
\usepackage{amsmath,amsfonts}
\usepackage{algorithmic}
\usepackage{algorithm}
\usepackage{array}
\usepackage[caption=false,font=normalsize,labelfont=sf,textfont=sf]{subfig}
\usepackage{textcomp}
\usepackage{url}
\usepackage{cite}
\usepackage{verbatim}
\usepackage{graphicx}
\usepackage{multirow}
\begin{document}
	
	\title{ Text-To-Image with Generative Adversarial Networks\\
		
		\author{\IEEEauthorblockN{Mehrshad Momen Tayefeh}
			\IEEEauthorblockA{{Sharif university of technology international campus-kish island}
				\textit{}
			}
	}}
	
	\maketitle
	
	\begin{abstract}
		Generating realistic images from human texts is one of the most challenging problems in the field of computer vision (CV). The meaning of descriptions given can be roughly reflected by existing text-to-image approaches. In this paper, our main purpose is to propose a brief comparison between five different methods base on the Generative Adversarial Networks (GAN) to make image from the text. In addition, each model architectures synthesis images with different resolution. Furthermore, the best and worst obtained resolutions is 64$\times$64, 256$\times$256 respectively. However, we checked and compared some metrics that introduce the accuracy of each model. Also, by doing this study, we found out the best model for this problem by comparing these different approaches essential metrics.  
	\end{abstract}

	
	\section{Introduction}
	Deep learning has transformed various fields, including classification \cite{liu2024atvitsc}, noise reduction \cite{10475264}, and even wireless communications through deep learning techniques \cite{MOMENTAYEFEH2025109710}. One of the most challenging tasks in deep learning is generating images from natural language descriptions. In recent years, several architectures have been proposed to tackle this challenge, aiming to generate highly detailed images from human-written captions by designing suitable deep learning models. However, most of these approaches are based on generative adversarial networks (GANs) \cite{goodfellow2020generative}. The quality of the generated images in GAN-based models depends on the interaction between the generator and discriminator.
	
	On the other hand, in text-to-image synthesis, there are lots of problems that can affect the model’s performance. In this field, unfortunately, there is no different dataset to generalize this task on every object. Also, the primary datasets are CUB-200-2011 \cite{wah2011caltech}, MSCOCO \cite{linmicrosoft}, and Oxford-102 flower \cite{nilsback2008automated}. In general, generating a real image from a human caption requires some vital features. For illustration, “generating an image from a general sentence requires multiple reasonings on various visual concepts, such as object category (people and elephants), spatial configurations of objects (riding), scene context (walking through a river), and etc.” \cite{hong2018inferring}.
	
	\begin{figure}[!t]
		\centering
		\includegraphics[width=3.4in,height=4.4in]{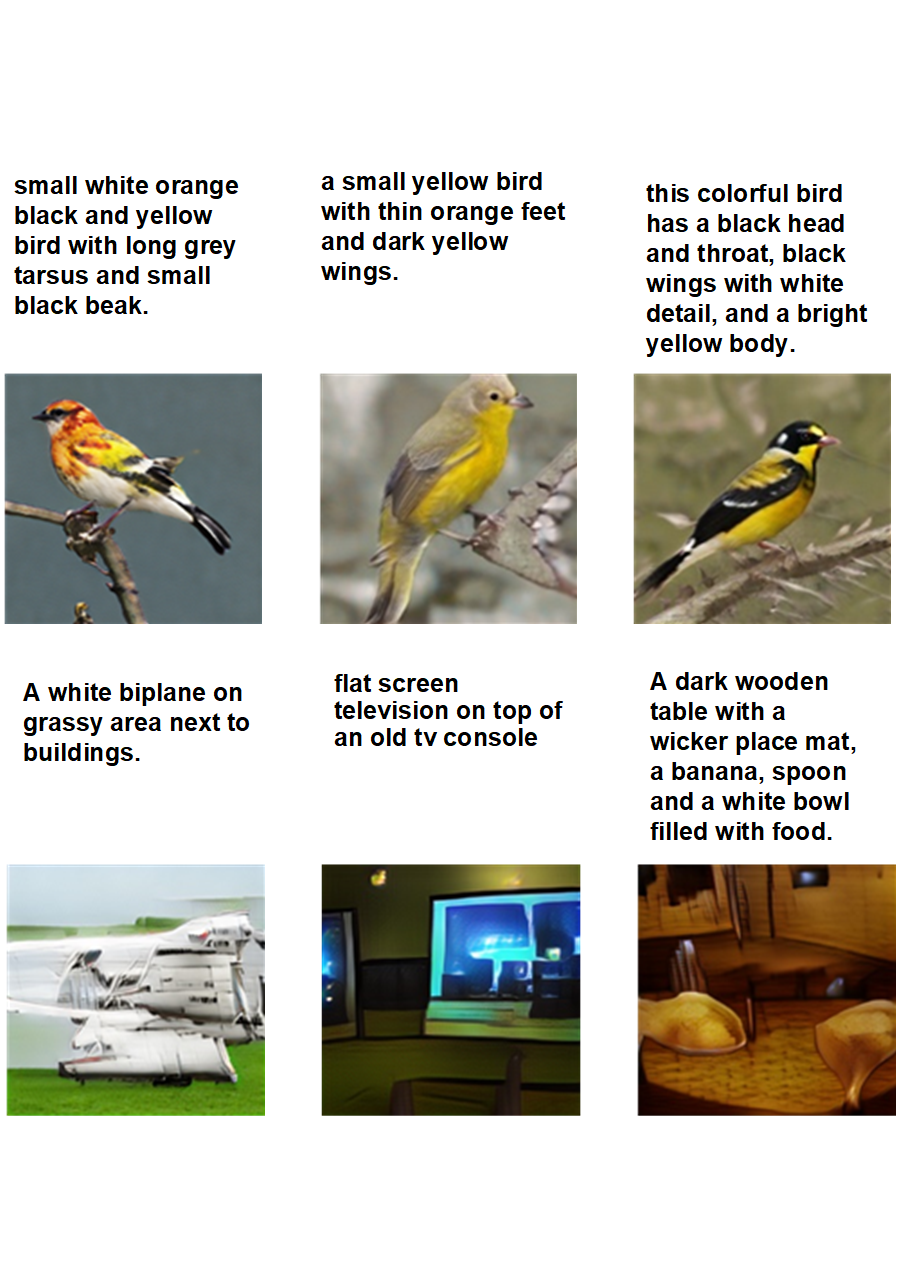}
		\caption{Example of different generated images from CUB-200 [2] and MSCOCO [3] dataset. First row images are from CUB-200-2011 and second are from MSCOCO dataset.}
		\label{drawing1}
	\end{figure}
	
	As mentioned, the main core of this task is the GAN model that contains two main networks, the generator, and the discriminator. In generator, the network takes a noise vector and converts it to an image. Therefore, the discriminator network takes the output of generator to check that its produced image is real and returns the result of comparison between the fake image of generator and the real image. 
	
	Figure \ref{drawing1} introduced an example of generated images from the models. Recently, there are proposed many different methods for text-to-image synthesis. Reed et al. \cite{reed2016generative} Introduce a basic and effective human text-to-image conversion model with deep convolutional  generative adversarial networks (DCGAN)\cite{radford2015unsupervised}. The article's authors \cite{dong2017semantic} present a symmetrical distillation network (SDN) that contains discriminative and generative networks as teachers and students, respectively. For the transformation of text into images, this paper proposes a two-stage model. Han Zhang et al. \cite{dong2017semantic} suggested a stackGAN that includes two-stage for generating text to high-resolution images. Hao dong et al. \cite{yuan2018text} present a model base on the conditional GAN architecture. In order to accurately synthesize photos from captions, Tao Xu et al. \cite{xu2018attngan} propose a new architecture based on attention mechanisms and combine it with DCGAN model.
	
	in this study, our main objective is to make a comparison between some ablation studies. In this task, we want to take a look at some plus and minuses in each different method and compare their performance. Additionally, we make a brief comparison between different models and methods. Also, the resolution and the naturalness of the photos in this study are so important.
	
	\begin{figure*}[h]
		\centering
		\includegraphics[width=\textwidth]{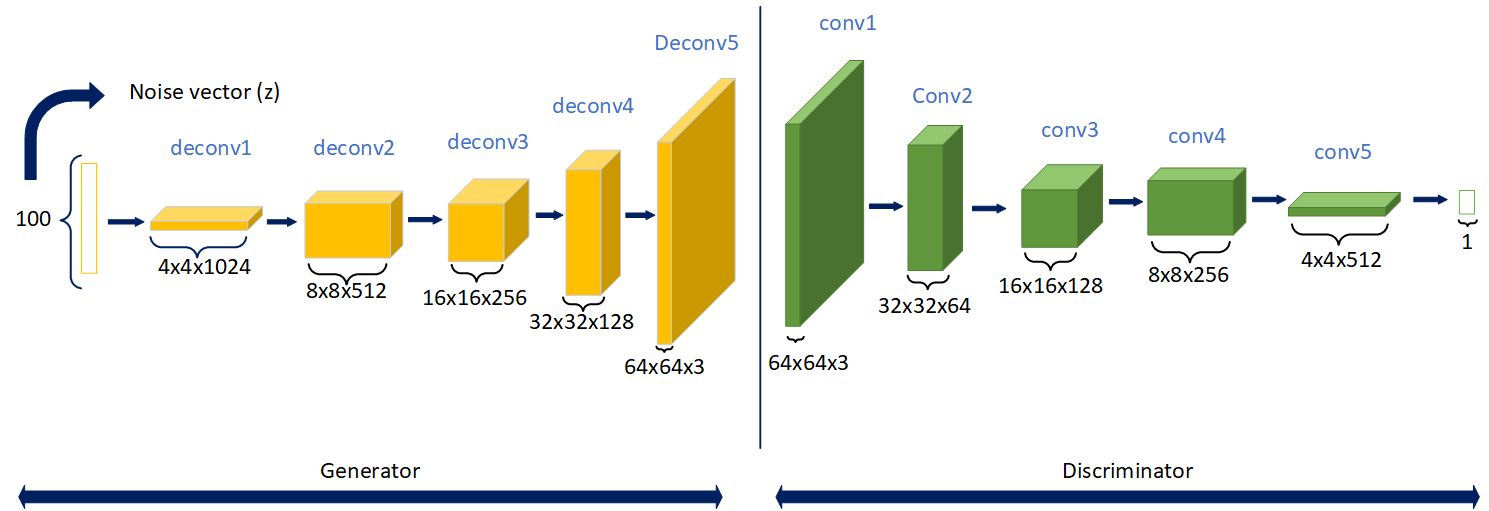}
		\caption{Simple architecture of a DCGAN that consist of 5 deconvolutional layers in generator and 5 convolutional layers in discriminator that generate a 64$\times$64$\times$3 image from noise vector with 100$\times$1 dimensions.  }
		\label{gan}
	\end{figure*}
	
	\section{Model Comparison}
	As mentioned in the previous part, our main objective is to propose a comparison between different parts of recent studies in the field of text-to-image synthesis. These similarities contain three main parts: first, the architecture of models used for this task. Second, datasets and evaluation metrics, and some other settings. In the last part, we check the contrast of output resolution.

	\subsection{Model Architectures}
	
	The most simple and regular model belongs to reed et al. \cite{reed2016generative} and is named that GAN-CLS. This architecture inspires DCGAN \cite{radford2015unsupervised} and comprises two main parts: Generator (G) and Discriminator (D) where Figure \ref{gan} is a representation of the DCGAN model for generate images. In the generator (G) first makes a noisy vector (z) that consists of random numbers coming from a uniform distribution; $z \in \mathbb{R}^Z \sim \mathcal{N}(0,1)$  and is sent as input to the network. Our discriminator (D) is just a classifier model that performs some convolutional layer with stride two and batch normalization \cite{ioffe2015batch} that followed by Relu as an activation function. Also, the output ‘0’ representation as fake and ‘1’ as an actual image. In general, GAN \cite{reed2016generative} network is like a minmax game; besides, Eq.1 is carried out GAN mechanism. In such a way, the generator (G) generating fake image try to reduce the output loss to produce more realistic images, and the discriminator (D), as the classifier, is in the role of referee. As well as, to compare the image generated by the generator (G) with the actual image and tries to increase the loss value. Additionally, after one iteration, generator (G) tries to make a better image. However, in GAN-CLS, we need to encode the text, images, and noise vectors. Image generation corresponds to feed-forward inference in the generator (G) conditioned on query text and a noise sample. In this model, our text encoder is Long-Short-Term-memory (LSTM) \cite{hochreiter1997long}, which extracts features of text and concatenates these with another part for the input generator.
	
	\begin{equation}
		\begin{aligned}
			\min_{G} \max_{D}  V(D,G)=\mathbb{E}_{x\sim p_{data}(X)}[\log D(X)]+ \dots \\ \mathbb{E}_{x\sim p_z(X)}[\log(1-D(G(z)))]
		\end{aligned}
	\end{equation} 
	
	Authors of \cite{dong2017semantic} use a baseline method that is a similar model to GAN-CLS but different in its generator network. Their model is constructed based on conditional GAN that conditions images and captions that describe it. The generator of this model consists of an encoder and a decoder. Encoders are employed to encode the main image and caption description. Decoders get the output features of encoders to synthesis images related to text descriptions. In addition, in this architecture, they adopt an MSE loss to train the encoder. The Eq.2 expresses this issue.
	
	\begin{equation}	
		\begin{aligned}
			loss_s = \mathbb{T}_{x\sim p_{data},z\sim p_z}\left\| z-S(G(z,\varphi(t)))) \right\|^2_2
		\end{aligned}
	\end{equation}
	
	On the other hand, two studies introduced novel models that employ different approaches. First, T. Xu et al. \cite{xu2018attngan} proposed a model that integrates the attention mechanism \cite{vaswani2017attention} with DCGAN, called the Attentional Generative Adversarial Network (AttnGAN). AttnGAN has two key components: the attentional generative network and the deep attentional multimodal similarity model (DAMSM). The final objective function of the attentional generative network is defined in Equation (3). The novel approach in this work uses a GAN-based model for text-to-image generation, with multiple generators ${(G_{m1}, G_{m2}, G_{m3}, \dots, G_{m-1})}$ that produce images using the same loss function (see Eq.3), which includes both the generator loss ($\mathcal{L}G$) and the DAMSM loss ($\mathcal{L}{DAMSM}$). In this equation, $\lambda$ represents a hyperparameter that balances the two terms in Eq.3. The number of discriminators matches the number of generators.

    The DAMSM consists of two submodules: a text encoder and an image encoder. The text encoder is a bidirectional LSTM that extracts features from captions, while the image encoder is a convolutional neural network (CNN) built on the inception-v3 model \cite{szegedy2016rethinking} pre-trained on ImageNet \cite{krizhevsky2012imagenet}. The image features are extracted from the ‘mixed-6e’ layer of the inception-v3 model. The main goal of AttnGAN is to synthesize high-resolution images using the final generator ${(G_{m-1})}$.
 
	\begin{equation}
		\begin{aligned}
			\mathcal{L}=\mathcal{L}_{G}+\lambda\mathcal{L}_{DAMSM},where\mathcal{L}_G=\sum_{i=0}^{m-1} \mathcal{L}_{G_i}
		\end{aligned}
	\end{equation}
	
	Second, H. Zhang et al. \cite{zhang2017stackgan} present an innovation architecture called StackGAN that carries out a two-stage model. In contrast of another architectures, this model firstly generates a low-resolution image at the stage-I. In this part focused on the general detail like proper colors and rough figures. For the generator $G_0$, to obtain text conditioning variable $c_0$, uses a Gaussian distribution $\mathcal{N} (\mu_{0}(\varphi_t), \sum_0(\varphi_t))$ that $\mu_0$ and $\varphi_0$ came from fed text embedding $\varphi_t$ into a fully connected layer and $c_0$ sampled from the Gaussian distribution. In stage-II, built upon the output of stage-I to generate the high-resolution image. Here try to fix the previous defect image to the most realistic image with all details. 
	
	Yuan et al. [9] proposed a symmetrical distillation network (SDN) that consists of the main target generator and main source discriminator as ‘teacher’ and ‘student’, respectively. The source discriminator in SDN is a common feature extraction like VGG19 \cite{simonyan2014very} model. This model has 16 convolutional layers with kernel size 3$\times$3 and three fully connected layers. In SDN model just used 16 convolutional layers for the discriminator. The generator in SDN has equal structures to the source discriminator. It consists of 3 fully connected layers and 16 convolutional layers again with kernel size 3$\times$3.
	
	\subsection{datasets and assessment metrics:}
	The datasets used in mentioned studies are CUB-200-2011 birds \cite{wah2011caltech} , Oxford-102 flowers [44], and Microsoft Common Objects in Context (MSCOCO) [3]. The CUB-200-2011 bird dataset contains 11,788 images from 200 different classes of birds; also, there is 10 caption per image that describes the image in detail. ‘’ Since 80\% of birds in this dataset have object-image size ratios of less than 0.5’’ [2]. The Oxford-102 has 102 classes of different flowers, and this dataset consists of 8,189 images, there is 10 caption that describe each image. The most challenging dataset is MSCOCO which is different from the noted datasets. The MSCOCO dataset is a large-scale object detection, segmentation, key-point detection, and captioning dataset. The dataset consists of 328K images with 80 different classes. It splits (83k) images into a training set, (41k) into a validation set, and (41k) into a testing set. In addition, for each image, define 5 captions related to the image.
	
	Moreover, \cite{reed2016generative,zhang2017stackgan} have trained their models on all mentioned datasets, and studies \cite{yuan2018text,dong2017semantic} apply and trained designed model on CUB-200-2011 and Oxford-102 datasets; Also, \cite{xu2018attngan} trained the AttnGAN on CUB-200-2011 and MSCOCO datasets.

	\section{Result and discussion}
	After comparing the model architectures, we will declare the results of the recent studies. The best performance metric for text-to-image synthesis is the Inception Score (IS) \cite{salimans2016improved}. The inception score is an objective performance metric for generated images from GAN models. It measures how realistic and diverse the output images are and is the second most crucial evaluation performance metric. This metric measures two things: first, Diversity, how diverse the generated images are. The entropy of the overall distribution should be high. Second, Quality how good the generated images are. Low entropy with high predictability is required. Therefore, a higher inception score is always better.
	
	By explaining the inception score, we compared this measurement for mentioned studies, in the \tablename.1. These results are related to AttnGAN, StackGAN, GAN-CLS, and SDN.
	
	\begin{figure*}[htb]
		\centering
		\includegraphics[width=\linewidth]{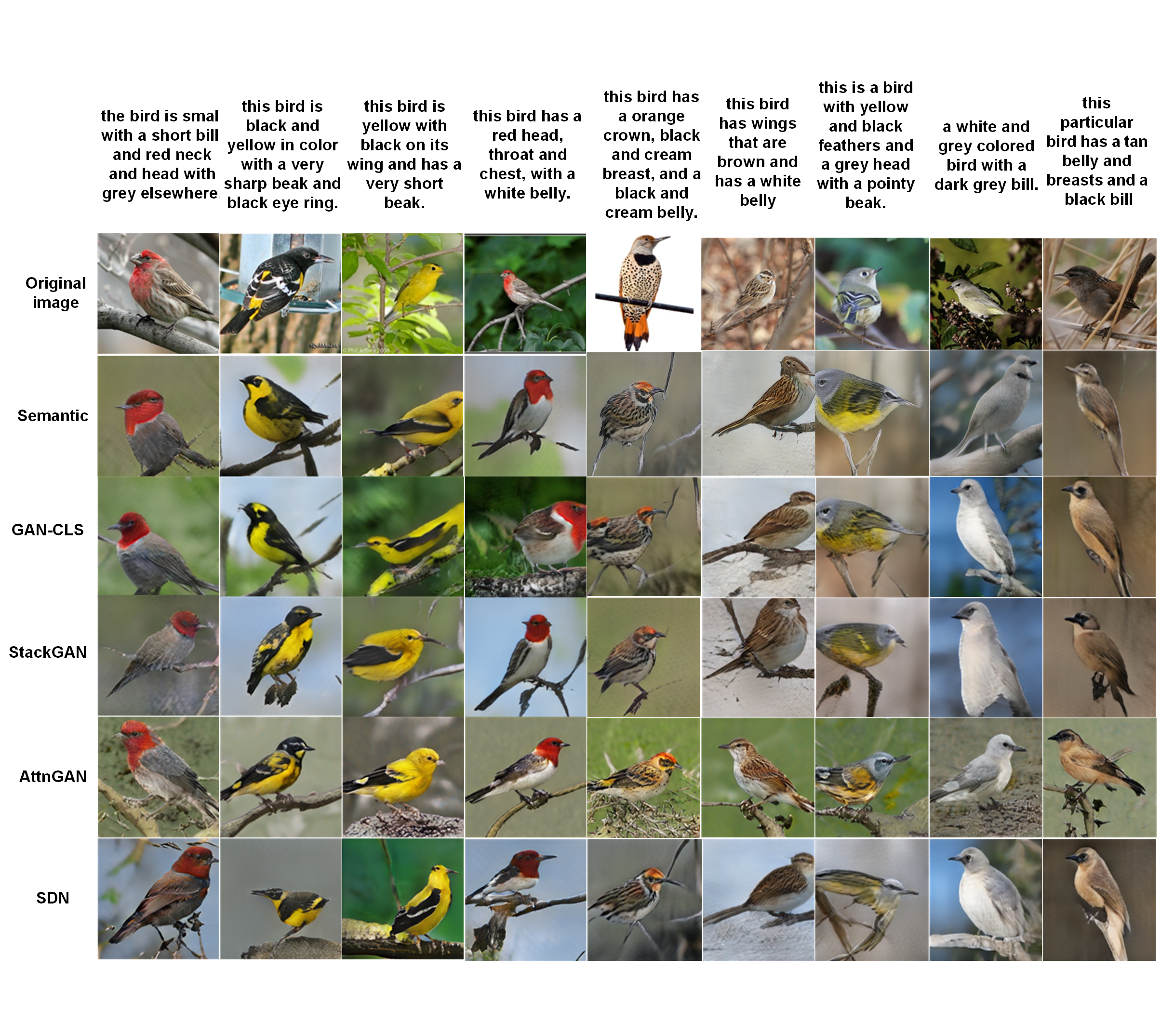}
		\caption{Example result of different images that generated from each mention methods on CUB-200-2011 dataset.}
		\label{fig:birds}
	\end{figure*}
	
	\begin{table}[htb]
		\centering
		\caption{Inception score for compared models on three primary datasets}
		\begin{tabular}{|p{0.14\linewidth}|p{0.15\linewidth}|p{0.15\linewidth}|p{0.16\linewidth}|p{0.15\linewidth}|}
			\hline
			\textbf{Dataset} & \textbf{AttnGAN \cite{xu2018attngan}} & \textbf{StackGAN \cite{zhang2017stackgan}} & \textbf{GAN-CLS \cite{reed2016generative}} & \textbf{SDN \cite{yuan2018text}} \\ 
			\hline
			\textbf{MSCOCO} & 25.89 $\pm$ .47 & 8.45 $\pm$ .03 & 7.88 $\pm$ .07 &  \\ 
			\hline
			\textbf{CUB-200-2011} & 4.36 $\pm$ .03 & 3.70 $\pm$ .04 & 2.88 $\pm$ .04 & 6.68 $\pm$ .06 \\ 
			\hline
			\textbf{Oxford-102} &  & 3.20 $\pm$ .01 & 2.66 $\pm$ .04 & 4.28 $\pm$ .09 \\ 
			\hline
		\end{tabular}
		\label{tab1}
	\end{table}
	
	By concentrating on the \ref{tab1}, due to all models did not use all mentioned datasets, we imply comparing by datasets. In other words, we try to find the highest IS for each dataset in referred models. Obviously, in the MSCOCO dataset, the AttnGAN had achieved the greatest IS with a considerable gap. After that, in the CUB-200-2011 and Oxford-102, the best IS was attained by the SDN model.
	
	On the other side, there is another metric for evaluating the models. For this reason, average human evaluation has been introduced. This metric is based on humans answering the output images. In other words, some humans give a score to how much the generated images are realistic. Each study randomly selects some text descriptions for every class on their datasets. After that, they will generate some images for each text and use human evaluation on each image. we have compared this appraisal in \ref{tab2}.
	
	\begin{table}[htb]
		
		\centering
		\caption{Average human evaluation for each referred dataset}
		\begin{tabular}{|p{0.13\linewidth}|p{0.15\linewidth}|p{0.15\linewidth}|p{0.16\linewidth}|p{0.15\linewidth}|}
			\hline
			\textbf{Dataset} &  StackGAN\cite{zhang2017stackgan} & GAN-CLS\cite{reed2016generative}& SDN\cite{yuan2018text}&Semantic with VGG\cite{dong2017semantic} \\
			\hline
			MSCOCO  &1.11 $\pm$ .03     & 1.89 $\pm$ .04 &&  \\ 
			\hline
			CUB-200-2011 & 1.37 $\pm$ .02     & 2.81 $\pm$ .03    & 2.26 $\pm$ 1.03  & 1.52     \\
			\hline
			Oxford-102  & 1.13 $\pm$ .03    & 1.87 $\pm$ .03    & 1.74 $\pm$ .77   & 1.49   \\
			\hline
		\end{tabular}
		\label{tab2}
		
	\end{table}
	
	The output size is much crucial in converting human texts to realistic images. Hence, compared with lower resolution to high-resolution images, more resolution can have to carry more details related to text description. In the text-to-image task, the most important part is the accuracy of the image and output resolution. Also, Figure \ref{fig:birds} demonstrated the result of image on bird dataset and here its visible the difference between models accuracy. The models of \cite{reed2016generative,dong2017semantic,yuan2018text} can synthesis images with 64$\times$64, 244$\times$244, and 224$\times$224, respectively. In contrast, the StackGAN and AttnGAN models are quite different. AttnGAN, due to having three outputs, therefor it generates images with the size 64$\times$64, 128$\times$128, and the final resolution is 256$\times$256. There are similarities for StackGAN. As said, this model consists of two stages; therefore, it has two output resolutions. The resolution in stage-I is 64$\times$64, and in stage II is 256$\times$256.

	By demonstrating this comparison between different models, our preferred model is AttnGAN. Due to the Attention mechanism in this model architecture, this proposed model has better performance and generates the most realistic image. The application of the Attention mechanism is in natural language processing—for instance, machine translation or in seq-to-seq models, inherent Attention. Therefore, using this mechanism in text-to-image synthesis is much helpful.
	
	\section{Conclusion}
	In this paper, different methods and models based on the Generative Adversarial Network (GAN) are compared for text-to-image synthesis. However, we have examined some necessary evaluation metrics for the best method. Recently used datasets for training the models were CUB-200-2011, Oxford102, and MSCOCO. The output image with high accuracy that completely described the input caption is the most important part of each model. By considering the objective of this study, due to the results, the AttnGAN achieved high inception score on the most challenging dataset (MSCOCO) and other datasets. Thereafter we can conclude that the introduced model is much more efficient.

	\bibliographystyle{IEEEtran}
	\bibliography{reference}
\end{document}